# P-trac Procedure: The Dispersion and Neutralization of Contrasts in Lexicon

Afshin Rahimi, Bahram Vazirnezhad, Moharram Eslami

**Abstract**
Cognitive acoustic cues have an important role in shaping the phonological structure of language as a means to optimal communication. In this paper we introduced P-trac procedure in order to track dispersion of contrasts in different contexts in lexicon. The results of applying P-trac procedure to the case of dispersion of contrasts in pre-consonantal contexts and in consonantal positions of CVCC sequences in Persian provide Evidence in favor of phonetic basis of dispersion argued by Licensing by Cue hypothesis and the Dispersion Theory of Contrast. The P-trac procedure is proved to be very effective in revealing the dispersion of contrasts in lexicon especially when comparing the dispersion of contrasts in different contexts.

1. **Introduction**

Traditionally, it has been argued that Phonological constraints account for phonological patterns. Recent studies on the role of articulatory/perceptual phonetic factors in phonological phenomena have provided consistent explanations for phonetically-based phonology (e.g. Flemming, 1995; Jun, 1995; Hamilton, 1996; Silverman, 1997; Steriade, 1997).

Licensing by Cue hypothesis, proposed by Steriade (Steriade, 1997), accounts for the role of phonetic factors in phonology. According to this hypothesis, maintenance of contrasts is closely related to the amount of perceptual salience a context can provide to a segment. The more the feature F of segment S is perceptually salient in context C, the more likely the S will show contrasts based on possible values of F in that context. Providing reasonable explanation for contrast neutralization, Licensing by Cue hypothesis has recently been challenged by some controversies. Kochetov argues that the distribution of Russian plain-palatalized contrast in coronal stops provides evidence against a complete phonetic-perceptual explanation claimed by Licensing by Cue hypothesis (Kochetov, 2006).

The Dispersion Theory of Contrast presents as an alternative explanations of dispersion/neutralization of contrasts in segment sequences (Flemming 1995, 2004, 2006). This hypothesis suggests three functional goals for phonological contrast dispersion.

- Maximization of the number of contrasts
- Maximization of perceptual distinctiveness of contrasts
- Minimization of articulatory effort for production of contrasts

Higher number of contrasts helps to distinguish words for efficient communication. The more the segments distinct perceptually, the easier they can be perceived by listeners. Finally minimization of articulatory effort accounts for efficiency of language production.

Functional goals accounted for phonological constraints have been reported in several works (Zipf, 1949; Martinet, 1952; Martinet, 1955; Lindblom, 1986; Lindblom, 1990), but there are arguments against functional basis of phonology (Ohala, 1993; Labov, 1994; Trask, 1996).

Licensing by Cue hypothesis and the Dispersion Theory of Contrast share the idea that perceptual factors play a prominent role in shaping phonological processes and patterns.

Consonant perceptual cues are good candidates for probing the role of phonetics in phonology and specially the role of perceptual cues in phonotactics. Consonants vary on the amount of contrast they reveal in different contexts. They show contrasts in prevocalic context but they are more limited in preconsonantal context. Sonority-based explanations, although make some predictions, but fail to provide a consistent prediction in all cases. Phonetically-based phonology, on the other hand, argues to provide such consistent predictions based on the assumption that phonotactics should ensure the perceptibility of cues to segmental contrasts (Kawasaki, 1982; Ohala, 1992; Flemming, 1995; Krichner, 1997; Steriade, 1999; Wright, 2004; Flemming, 2007). The studies show evidence against a sonority-based account of phonotactics and make strong arguments on the role of perceptual cues to segmental contrast in phonotactics of the world languages.

Based largely on Wright (2004), a number of cues to place, manner and voicing are introduced here. Perceptual cues to place contrasts are related to acoustic features including second formant transitions, stop release bursts, nasal pole zero patterns and fricative noise. These features are more recoverable in the short period of transition to the next segment especially in the case of perceptual cues to stops place contrasts. Fricatives have cues internal to the fricative signal itself. Some laterals have cues spreading over an entire syllable (Wright, 2004). Due to the more susceptibly of stop place contrasts to cue loss especially in noisy channels; Languages put more restrictions on the position of stops in segment sequences.

The manner contrasts are recoverable by relative degree of attenuation in the signal as a perceptual cue. An abrupt attenuation in signal is a sign of a stop. A complete attenuation along with fricative noise is a manner cue to fricatives. Nasals have a small decrease in amplitude of the signal compared to fricatives. They also use nasal pole and zero as a cue to nasal manner. Manner cues are more resistant to noise masking and are more perceptually salient compared to voice and place cues (Wright, 2004).

Cues to voicing are periodicity, VOT, the presence and the amplitude of aspiration noise, and duration cues which can be irrecoverable for stops in syllable final and preconsonantal contexts (Wright, 2004).

The study of perceptual cues has revealed that cues to voicing contrasts are weaker than cues to place contrasts and cues to place contrasts are weaker than cues to manner contrasts especially in preconsonantal contexts (Wright, 2004). According to Licensing by Cue hypothesis if perceptual cues of a contrast are weak in a context it is less likely for a segment to show contrast of the feature in that context and the contrast will be subject to neutralization.

The aim of this study was to investigate Licensing by Cue hypothesis (Steriade, 1997) in the case of contrast neutralization of consonants in preconsonantal context. Segments are predicted to show more contrast according to different values of features more perceptually salient in preconsonantal context. Voicing contrasts are weaker than place contrast and place contrasts are weaker than manner contrasts, thus if the hypothesis is true, consonants in preconsonantal context should show the most contrasts on the manner dimension, a medial amount of contrast on the place dimension and the least amount of contrasts on the voicing dimension.

Another goal of this paper is to introduce P-trac procedure, a procedure to find the distribution of contrasts in a context within lexicon. The assumption is if a cue is weak in a context, the contrasts related to that cue will be subjected to neutralization. Diachronically the amount of contrasts in the poor contexts will be reduced while the amount of contrasts of perceptually salient cues will be increased in the proper contexts within lexicon. A procedure can track the distribution of contrasts in a context and provide insights into the degree of perceptibility of a cue in that context.

## 2. P-trac Procedure

In this section, we will introduce the P-trac procedure. The goal of P-trac is to track the dispersion of contrasting features in lexicon. The assumption behind the P-trac procedure is that if phonology is perceptually grounded, the lexicon should be diachronically optimized so that enough cues for each contrasting feature exist in segmental contexts. As an example, cues to voicing contrasts are weaker in preconsonantal contexts than in prevocalic contexts so the amount of voicing contrast should be less frequent in preconsonantal contexts than in prevocalic contexts within a lexicon according to diachronic phonological changes. To find the dispersion of contrasting features, we need to define the notion of featural minimal pair and minimal sequence pair. A featural minimal pair is a pair of segments which are the same in all features except one contrasting feature. Segments /b/ and /p/ constitute a featural minimal pair because they share the same features except voicing feature while /b/ and /t/ don't constitute a featural minimal pair because they contrast both in voicing feature and place feature. A minimal sequence pair is defined as two sequences of segments which differ only in segments of one position while the two segments should be themselves featural minimal pairs. For example /band/ and /dand/ are a minimal sequence pair because they only differ in segments /d/ and /b/ in the starting position of the sequence and segments /b/ and /d/ are themselves a featural minimal pair which differ only in place feature. On the other hand /band/ and /tand/ don't form a minimal sequence pair because /b/ and /t/ don't constitute a featural minimal pair. The notion of featural minimal pair is different from the notion of minimal pair usually used in phonology. Two segment sequences that differ in only one phoneme and have distinct meanings are called a minimal pair in phonology literature. Minimal pairs are used to construct phoneme inventory of a language while featural minimal pairs are used to distinguish a contrasting feature. Table-1 shows the possible featural minimal pairs and their contrasting feature in Persian.

| Contrastive Feature | Number of featural minimal pairs | Featural Minimal Pairs | Acoustic Cues |
|---|---|---|---|
| Manner | 70 | (b, m) (b, w) (b, v) (C, t) (C, s) (C, S) (d, n) (d, r) (d, z) (d, l) (d, j) (d, Z) (f, p) (g, y) (j, n) (j, r) (j, z) (j, l) (j, d) (j, Z) (k, x) (l, z) (l, r) (l, d) (l, n) (l, Z) (l, j) (m, w) (m, b) (m, v) (n, z) (n, r) (n, l) (n, d) (n, j) (n, Z) (p, f) (q, y) (r, d) (r, z) (r, j) (r, n) (r, Z) (r, l) (s, t) (s, C) (S, t) (S, C) (t, s) (t, S) (t, C) (v, b) (v, m) (v, w) (w, b) (w, v) (w, m) (x, k) (y, g) (y, q) (z, r) (z, n) (z, l) (z, d) (z, j) (Z, r) (Z, n) (Z, l) (Z, d) (Z, j) | degree of signal attenuation |
| Place | 50 | (', b) (', q) (', d) (', g) (b, ') (b, q) (b, d) (b, g) (d, q) (d, ') (d, b) (d, g) (f, s) (f, S) (f, x) (f, h) (g, ') (g, d) (g, b) (h, s) (h, x) (h, f) (h, S) (k, t) (k, p) (m, n) (n, m) (p, t) (p, k) (q, d) (q, b) (q, ') (s, f) (s, x) (s, h) (S, x) (S, f) (S, h) (t, k) (t, p) (v, z) (v, Z) (w, y) (x, s) (x, f) (x, S) (x, h) (y, w) (z, v) (Z, v) | formant transitions stop release bursts nasal pole & zero fricative noise |
| Voice | 20 | (b, p) (C, j) (d, t) (f, v) (g, k) (j, C) (k, q) (k, g) (p, b) (q, k) (s, z) (s, Z) (S, Z) (S, z) (t, d) (v, f) (z, S) (z, s) (Z, S) (Z, s) | periodicity VOT aspiration noise duration |

Table-1: All possible featural minimal pairs in Persian, their contrastive feature and their corresponding acoustic cues

P-trac procedure starts with extracting minimal sequence pairs from lexicon. Minimal sequence pairs should be extracted according to the subject and goals of the study. As an example in order to study the dispersion of manner, place and voice features in different context in CVCC syllables of Persian, it required to find and extract all minimal sequence pairs of CVCC type syllables from a Persian lexicon. In another case if the goal of study is to find the distribution of consonant features in preconsonantal context of CVCC syllables of Persian, All minimal sequence pairs in the form of $C_1C_2$ from CV$C_1C_2$ syllables should be extracted. The basic idea of P-trac procedure is given in (1).

(1)

a) Extract all sequences according to the goals of study.
b) Find all minimal sequence pairs.
c) For each minimal pair designate context and the contrasting feature.
d) Count the frequency of occurrence for each pair *(context, contrasting feature)*.

At the end of the P-trac procedure we'll have a feature-context matrix that involves the frequency of occurrence for all contrasting features in all contexts. A general feature-context matrix is shown in (2).

(2)

|  | $Context_1$ | $Context_2$ | $Context_i$ |
|---|---|---|---|
| **Place** | $freq$ | $freq$ | … |
| **Manner** | $freq$ | $freq$ | … |
| **Voice** | $freq$ | $freq$ | … |

P-trac procedure can be applied to find the distribution of one or several contrasting features in one or several contexts. For example, it can be used to find the dispersion of all contrasting features before the voiced labial stop segment /b/ in $C_1$b consonant clusters (a single context) or it may be used to find the distribution of voicing contrast (a single contrasting feature in all preconsonantal contexts). Similar algorithms to the P-trac procedure have been proposed in the literature in order to find contrasting features but they are used for other goals other than tracking the perceptual optimization of a lexicon (e.g. Archangeli, 1988; Dresher, 2003).

3. **Using P-trac procedure to find the distribution of contrasting features in preconsonantal context of CVCC segment sequences in Persian**

In this experiment, we examined the distribution of consonantal contrasts in preconsonantal context. The goal of this experiment was to know whether the distribution of contrasts matches the amount of perceptibility of cues. According to Wright (2004), the perceptibility of cues decreases from cues to manner contrasts to that of place contrasts and from cues to place contrasts to that of voicing contrasts. If Licensing by Cue hypothesis (Steriade, 1997) holds, due to its diachronic effect, we should see a correlation between the amount of the perceptibility of the acoustic cues to features and the number of times the feature is used to contrast a minimal pair.

3.1. **Data**

Just like CELEX lexicon (Baayen et al., 1995) for some western languages, FLexicon is a database which contains information about the lexicon of Persian language. This database contains 54409 words and their phonemic transcription. Syllable structure in Persian is very simple because it doesn't allow complex onsets. The only three possible syllable structures are CV, CVC and CVCC. Therefore, the syllabification of lexemes can be done deterministically.

3.2. **Applying P-trac procedure**

According to the P-trac procedure, we syllabified all 54409 lexemes in lexicon using simple rules of Persian syllabification. In Persian, syllables should start with an obligatory Onset. Moreover, Onset clusters are forbidden in Persian so we can simply syllabify lexemes. Each syllable in Persian starts with a consonant as Onset followed by a vowel. In spite of the Maximal Onset Principle which is used to syllabify segment sequences in English, a so called Minimal Onset Principle is used to syllabify sequences in Persian according to which all consonant clusters belong to coda except the last consonant which is an obligatory onset. For example, CVCCCV sequence is always syllabified as CVCC.CV because there must be just one obligatory consonant as the Onset of the second syllable. In a similar way, CVCV sequence is always syllabified as CV.CV. So the syllabification of Persian segment sequences is a simple deterministic task.

After syllabification of lexemes, 268 distinct $C_1C_2$ clusters were extracted from CV<u>CC</u> syllables and their frequencies were counted within the lexicon (type frequency).

For each $C_2$ in consonant cluster $C_1C_2$, we found all featural minimal pairs in $C_1$ position. We defined a featural minimal pair as two phonemes distinguished by just one contrastive feature. The features we used were manner, place and voice.

The distribution of contrastive features was investigated by counting the number of times each feature used to form a minimal consonant pair in $C_1$ position in preconsonantal context ($\_C_2$) within $C_1C_2$ consonant clusters. For example in all $C_1r$ consonant clusters, all minimal pairs were extracted and the contrasting feature was counted. For each feature (manner, place, voice), a frequency of occurrence in context $\_C_2$ (here /_r/) was resulted. The procedure is described in (3).

(3)

> for each consonant in $C_2$ position of $C_1C_2$ consonant cluster
> > for each $C_{11}$ in $C_1$ position
> > > for each $C_{12}$ in $C_1$ position
> > > > If $C_{11}$ and $C_{12}$ are featural minimal pair
> > > > > feature ← Contrastive Feature of $C_{11}$ and $C_{12}$
> > > > > freq ← min(frequency of $C_{11}C_2$, frequency of $C_{12}C_2$)
> > > > > frequencies[$C_2$, feature] = freq + frequencies[$C_2$, feature]

At the end of the procedure, for each consonant in $C_2$ position (context) and for each feature (manner, place and voice) the number of times the feature was used as the contrasting feature in a featural minimal pair in $C_1$ position is counted.

### 3.3. Results

Diagram-1 demonstrates the result of applying P-trac procedure applied to $C_1C_2$ clusters in CVCC syllables extracted from FLexicon. The consonants in $C_2$ position provide the preconsonantal context for $C_1$ consonants. It visualizes the distribution of contrasting features which consonants in $C_1$ position use to distinguish meaning in preconsonantal context $\_C_2$. There are a total of 791 minimal sequence pairs from which 433 pairs contrast in manner, 335 pairs contrast in place and 23 pairs contrast in voice feature. It should be noted that according to P-trac procedure given in (3), the frequencies are computed using the minimum type frequency of each sequence in lexicon. For example, if the type frequency of /bl/ is 200 (it means there are 200 /bl/ sequences in all distinct CVCC sequences in the lexicon) and that of /pl/ is 300 in the lexicon we increment *frequencies[l, voice]* by 200, the minimum type frequency of the two sequences.

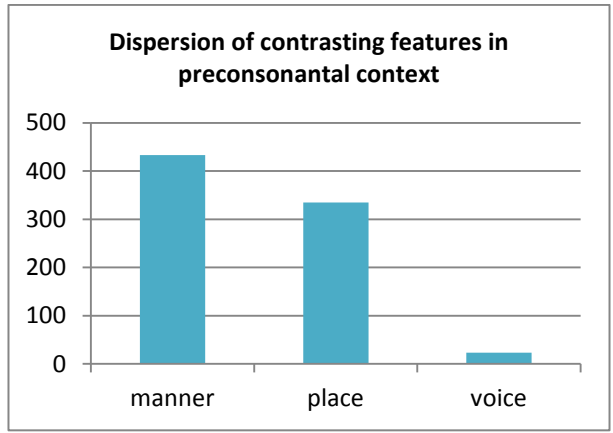

Diagram-1: Frequency distribution of contrasting features in $C_1$ position of $CVC_1C_2$ sequences

Although the results in diagram-1 are summed over all the consonants in $C_2$ position, P-trac procedure provides the dispersion of the contrasts for all the consonant instances. We used the output of P-trac procedure in order to find in which contexts consonants in $C_1$ position use voicing as a contrasting feature because perceptual cues to voicing contrasts are weak in preconsonantal contexts. Diagram-2 demonstrates all the contexts in which voicing feature has been used as the contrasting feature in order to distinguish meaning in the FLexicon, the Persian lexicon. As shown in the diagram, the only contexts that voicing contrast is used are those before nasals and liquids. In other preconsonantal contexts, voicing contrast is not used anywhere.

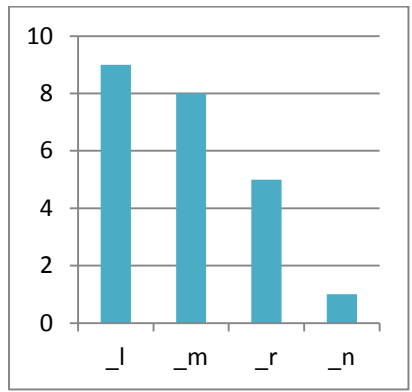

Diagram-2: Preconsonantal Contexts in $CVC_1C_2$ sequences where voicing is used as a contrasting feature

Table-2 demonstrates the minimal sequence pairs that use voicing contrast in order to distinguish meaning in the FLexicon. The surprising fact about these minimal sequence pairs is that all of them are either loan words or part of loan words all borrowed from Arabic language.

| Preconsonantal Context | Number of Contrasts | Minimal Sequence Pairs |
|---|---|---|
| _l | 9 | (/ʔasl/, /ʔazl/) (/fasl/, /fazl/) |
| _m | 8 | (/rasm/,/razm/) |
|  | 5 | (/ʔosr/, /ʔozr/) |

|  |  | (/ʔoSr/,(/ʔozr)  |
| --- | --- | --- |
| _r |  | (/nasr/, /nazr/) |
|  |  | (/naSr/, /nazr/) |
|  |  | (/satr/, /sadr/) |
| _n | 1 | (/hosn/, /hozn/) |

**Table-2: Minimal sequence pairs that use voicing contrast in Persian lexicon**

## 4. Using P-trac procedure to compare the dispersion of contrasting features in consonantal positions of CVCC syllables in Persian.

In this experiment, the P-trac procedure was used in order to find the dispersion of manner, place and voicing features in $C_1, C_2$ and $C_3$ positions of $C_1VC_2C_3$ syllables in FLexicon. The goal of this experiment is to find how distribution of contrasting features is related to the amount of the perceptibility of the cues argued by Licensing Cue hypothesis (Steriade, 1997) and Dispersion Theory of Contrast (Flemming 1995, 2004, 2006). The Phonemic transcriptions of lexemes were used again as the input to P-trac procedure.

### 4.1. Applying P-trac procedure

All distinct CVCC syllables with their type frequency, the frequency of the syllable in the lexicon, were extracted. According to P-trac procedure, all minimal sequence pairs were found and for each context (consonants in $C_1, C_2$ and $C_3$ position of $C_1VC_2C_3$) the contrasting features were counted. Just like the previous experiment for each minimal sequence pair, the *frequencies[context, feature]* was added by the minimum type frequency of members of the minimal sequence pair. For example, for /band/ and /pand/ minimal sequence pair if the type frequency of /band/ is 200 and that of /pand/ is 300, *frequencies[_and, voice]* was added by 200.

### 4.2. Results

In diagram-3 the distribution of manner, place and voicing contrasts are shown. As it can be seen the voicing contrast is minimal in preconsonantal context $C_2$ and is maximal in onset position (prevocalic context $C_1V$).

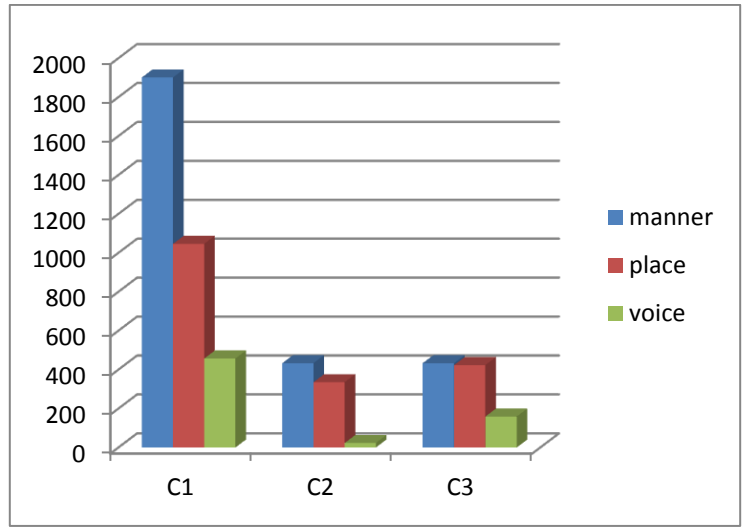

**Diagram-3:** Frequency distribution of contrasting features in C1, C2 and C3 positions in C1VC2C3 syllables of Persian lexicon.

As it can be seen the frequency of contrasts is maximal in prevocalic context $C_1$ (3404 contrasts), medial in position $C_3$ (1015 contrasts) and minimal in preconsonantal context $C_2$ (791 contrasts) whatever the contrasting feature is. The frequency of contrasting features gradually decreases from manner feature to place feature and from place feature to voicing feature.

## 5. Discussion

The results of P-trac procedure applied to the case of contrasting features in preconsonantal contexts of $C_1C_2$ clusters in CVCC syllables extracted from FLexicon shows that the dispersion of contrasting features exactly matches with the predictions made by Licensing by Cue hypothesis and the Dispersion Theory of Contrast. The frequency of contrasting features in preconsonantal context gradually decreases from manner contrasts to place contrasts and from place contrasts to voicing contrasts (Diagram-1). According to (Wright, 2004) the amount of perceptibility of acoustic cues of manner, place and voicing features has exactly the same pattern. The more salient is the perceptual cue of a feature in preconsonantal context, the more frequently it has been used as a contrasting feature to distinguish meaning. This statistical evidence, the result of P-trac procedure, provides support for phonetic basis of phonology in general and Licensing by Cue hypothesis and the Dispersion Theory of Contrast hypothesis in particular.

Voicing contrast has the least salient perceptual cues in preconsonantal contextscompared to manner and place contrasts. The results show that from 793 minimal sequence pairs only 23 pairs have used voicing as the contrasting feature. The output of P-trac procedure shows that the only context that voicing contrast has been used is pre-sonorant consonants. Voicing contrast isonly used before liquids and nasals (Diagram-2). This provides support for the hypothesis that voicing contrasts has more salient perceptual cues before liquids and nasals. In other contexts such as before an obstruent the voicing contrast neutralizes so the voicing feature is not used as a contrasting feature before stops and fricatives.

Another fact revealed by the output of P-trac procedure is that all words that make use of voicing feature as the contrasting feature in preconsonantal contexts are loan words from Arabic language (Table-2). Persian has borrowed many words from Arabic language and the interestingly all the voicing contrasts in preconsonantal contexts are related to those loan words.

The results of applying P-trac procedure on $C_1VC_2C_3$ syllable sequences of Persian shows that the frequency of contrasts decreases from onset position $C_1$ to position $C_3$ and from $C_3$ to position $C_2$. The distribution of contrasts in the 3 contexts is again exactly matched with the amount of perceptibility of cues to features in those contexts. The perceptibility of cues is maximal in onset position, medial in last coda position and minimal in preconsonantal position (Wright, 2004) and has the exact pattern of the dispersion of contrasts in those contexts. This again provides support for the direct relation between amount of perceptual salience of cues to features in a context and amount of contrasts in that context.

## 6. Conclusion

In this paper we introduced P-trac procedure in order to track dispersion of contrasts in different contexts in lexicon. The results of applying P-trac procedure to the case of dispersion of contrasts in preconsonantal contexts and in consonantal positions of CVCC sequences in Persian provide Evidence in favor of phonetic basis of the dispersion of contrasts argued by Licensing by Cue hypothesis and the Dispersion Theory of Contrast. The P-trac procedure is proved to be very effective in revealing the dispersion of contrasts in lexicon especially when it provides means to compare the dispersion of contrasts in different contexts.